\documentclass[manuscript,screen, anonymous=false]{acmart}
\usepackage{array}
\usepackage[capitalize]{cleveref}

\AtBeginDocument{%
  }

\setcopyright{acmlicensed}
\copyrightyear{2024}
\acmYear{2024}
\acmDOI{XXXXXXX.XXXXXXX}

\acmConference[Conference acronym 'XX]{Make sure to enter the correct
  conference title from your rights confirmation emai}{June 03--05,
  2018}{Woodstock, NY}

\usepackage[capitalize]{cleveref}
\usepackage{longtable}
\usepackage{comment}
\begin{document}


\title{More than Marketing? On the Information Value of AI Benchmarks for Practitioners}



\author{Amelia Hardy}
\authornote{Both authors contributed equally to this research.}
\email{amelia@cs.stanford.edu}

\author{Anka Reuel}
\authornotemark[1]
\email{anka@cs.stanford.edu}
\affiliation{%
    \institution{Stanford University}
  \city{Stanford}
  \state{California}
  \country{USA}
}

\author{Kiana Jafari Meimandi}
\affiliation{%
    \institution{Stanford University}
  \city{Stanford}
  \state{California}
  \country{USA}
}
\email{kjafari@stanford.edu}

\author{Lisa Soder}
\affiliation{%
  \institution{interface – Tech Analysis and Policy Ideas for Europe e.V.}
  \city{London}
  \country{United Kingdom}
}
\email{lsoder@interface-eu.org}

\author{Allie Griffith}
\affiliation{%
    \institution{Stanford University}
  \city{Stanford}
  \state{California}
  \country{USA}
}
\email{allie11@stanford.edu}

\author{Dylan M. Asmar}
\affiliation{%
    \institution{Stanford University}
  \city{Stanford}
  \state{California}
  \country{USA}
}
\email{asmar@stanford.edu}

\author{Sanmi Koyejo}
\affiliation{%
  \institution{Stanford University}
  \city{Stanford}
  \state{California}
  \country{USA}
  } 
\email{sanmi@stanford.edu}

\author{Michael S. Bernstein}
\affiliation{%
  \institution{Stanford University}
  \city{Stanford}
  \state{California}
  \country{USA}
  }
  
\email{mbs@cs.stanford.edu}

\author{Mykel J. Kochenderfer}
\affiliation{%
  \institution{Stanford University}
  \city{Stanford}
  \state{California}
  \country{USA}
  }
  
\email{mykel@stanford.edu}
\renewcommand{\shortauthors}{Hardy et al.}

\begin{abstract}
Public AI benchmark results are widely broadcast by model developers as indicators of model quality within a growing and competitive market. However, these advertised scores do not necessarily reflect the traits of interest to those who will ultimately apply AI models. In this paper, we seek to understand if and how AI benchmarks are used to inform decision-making. Based on the analyses of interviews with 19 individuals who have used, or decided against using, benchmarks in their day-to-day work, we find that across these settings, participants use benchmarks as a signal of relative performance difference between models. However, whether this signal was considered a definitive sign of model superiority, sufficient for downstream decisions, varied. In academia, public benchmarks were generally viewed as suitable measures for capturing research progress. By contrast, in both product and policy, benchmarks -- even those developed internally for specific tasks -- were often found to be inadequate for informing substantive decisions. Of the benchmarks deemed unsatisfactory, respondents reported that their goals were neither well-defined nor reflective of real-world use. Based on the study results, we conclude that effective benchmarks should provide meaningful, real-world evaluations, incorporate domain expertise, and maintain transparency in scope and goals. They must capture diverse, task-relevant capabilities, be challenging enough to avoid quick saturation, and account for trade-offs in model performance rather than relying on a single score. Additionally, proprietary data collection and contamination prevention are critical for producing reliable and actionable results.  By adhering to these criteria, benchmarks can move beyond mere marketing tricks into robust evaluative frameworks that accurately reflect AI progress and guide informed decision-making in both research and practical domains. 

\end{abstract}



\received{20 February 2007}
\received[revised]{12 March 2009}
\received[accepted]{5 June 2009}

\maketitle

\section{Introduction}
\label{sec:introduction}
When choosing between different AI models, users are faced with a growing number of increasingly capable options \citep{maslej2024ai}. Such decisions may be informed by model evaluations, which commonly include manual assessment, red-teaming, and the use of AI benchmarks~\citet{chen2021evaluating}. Benchmarks, which following \citet{raji2021benchmark} we define ``as a particular combination of a dataset or sets of datasets [...], and a metric, conceptualized as representing one or more specific tasks or sets of abilities, picked up by a community of researchers as a shared framework for the comparison of methods'', are frequently cited in publications and reports from model developer  \cite{tascilar2023quest,bender2021dangers}. Although such reports frame scores on public benchmarks as the basis one should have for understanding a model's capabilities, it is not evident that this framing is consistent with how benchmarks are used in practice. Despite what developers who publish their scores may suggest, benchmark results may not give model users enough information to make substantive, final choices about which model is adequate for their use case. 

In this work, we seek to understand how AI benchmarks inform decisions between models. To build our theory, we perform and analyze semi-structured interviews with 19 practitioners in academia, policy, and industry, including research scientists, PhD students, and machine learning engineers. The study aims to address the following core research questions:
\begin{itemize}
    \item How are AI benchmarks developed, assessed, and used, across different application areas?
    \item What challenges do stakeholders face in using and interpreting these benchmarks?
    \item To what extent do AI benchmarks impact decision-making and progress in research, product, and policy?
\end{itemize}

Our analysis reveals that in the domains of research, product, and policy, individuals are comfortable using benchmarks to make relative comparisons between models. These comparisons more frequently serve negative decisions more than positive ones: a low benchmark score can stop a model from being deployed, but a high score does not guarantee deployment. When it comes to real-world usage decisions, product and policy practitioners often find existing benchmarks to be insufficient. The ability of existing benchmarks to facilitate comparison does not obviate these practitioners' need for further assessment, leading them to either develop internal benchmarks or to use alternate forms of evaluation. This contrasts with how benchmarks function in research settings, where the mere existence of a benchmark can drive growth within a sub-field. Thus, the question remains: what do benchmarks measure progress towards?


We propose that the openness of this question is a key factor impacting the adoption of AI benchmarks. Drawing from theories of information technology acceptance \citep{davis1986tam}, we find that although benchmarks meet the majority of the criteria for technology adoption, they fall short on the factor of performance expectancy, i.e., a technology's perceived ability to perform the task it is assigned. A recurring theme among participants in our study was that of a gap between the task a benchmark tests a model on and how that model will ultimately be used. While there are fundamental limits to how well a test or simulation task can approximate real-world applications \cite{kadian2020sim2real}, we found that participants were able to overcome part of the gap by developing internal evaluations to better reflect their actual use cases. We find that benchmark developers can make their benchmarks more informative and useful to practitioners by considering user personas and use cases during the design process, incorporating feedback from domain experts, and clearly documenting how results should and should not be interpreted. 


\section{Related Work}
\label{sec:related_work}
Our literature review considers three broad domains as the basis for understanding how AI benchmarks are (and are not) used to support decisions. First, to position our findings, we review the literature on the recognized issues with current AI evaluation practices. Although some issues with model assessment are benchmark-specific, others remain open questions in the field. Second, to contextualize current AI benchmarks, we review literature on the history of benchmarking and analyze the motivations that have driven past benchmark use in computing and other fields. Finally, to map the circumstances under which benchmarks may be adopted, we consider models for the acceptance of information technology.

\subsection{Evaluating AI}
\label{sec:eval_ai}
Evaluating AI systems is necessary for understanding their capabilities, limitations, and potential for positive and negative impacts~\cite{saxon2024benchmarks}. However, this process is complex, requiring careful consideration of which evaluation methods are used and how their results are interpreted. 

One of the primary challenges in AI evaluation is ensuring the closeness of an evaluation to a system's anticipated use~\cite{raji2021benchmark,beede2020human}. This issue is exemplified by a study conducted by~\cite{lebovitz2021ground}, which examined five different AI tools deployed in a U.S. hospital. The study found that none of these tools met expectations created by accuracy measured using expert-labeled ground truth data. This highlights the risk of treating constructed quality measures as objective markers of knowledge and underscores the importance of diligent, use-case specific expert evaluation~\cite{lebovitz2021ground}. This phenomenon of metrics used to evaluate AI models lacking correlation with users' subjective experiences has been documented across applications. \citet{gordon2021disagree} note that models which score highly on standardized tests have been observed and criticized for their obvious mistakes. While bridging the gap between assessment and reality is an inherent issue with AI evaluation \citep{kadian2020sim2real}, we note that it is possible to mitigate the issue to some extent through careful, case-specific design \citep{grabb2024risks, reuel2024betterbench}.

Another challenge in AI evaluation is the inherent difficulty of clearly defining the capabilities AI systems are supposed to exhibit and, subsequently, ensuring construct validity \cite{cronbach1955construct} with respect to the test tasks. While performance on certain tasks like image classification or speech recognition can be clearly-defined, more complex tasks - such as decision-making under uncertainty or ethical reasoning - are much harder to pin down~\cite{bostrom2020public}. Since AI systems often operate in environments that are unpredictable and involve nuanced human behaviors, it is difficult to set clear, measurable objectives for evaluation~\cite{floridi2024three}. In a study of the challenges faced by software engineers developing AI systems, \citet{patel2008examining} find that evaluating performance is a significant challenge, in part due to the trade offs between different attributes, e.g., accuracy and the protection of privacy. That the former is more easily captured than the latter does not make it a better optimization objective.

Further complicating the issue of assessment, AI systems tend to excel in narrow, known tasks, while struggling to generalize to new, unseen scenarios. Evaluations struggle to capture this property, as they are typically based on datasets that represent a small slice of reality. As a result, evaluations can suffer from overfitting, wherein systems perform well on dataset that resemble their training data but fail to generalize when tested on others. This challenge exposes the limitations of controlled, static testing environments that cannot capture the full variability of real-world situations~\cite{gebru2021datasheets}. The failure of AI systems to generalize well in diverse and dynamic settings underscores the difficulty in designing evaluations that meaningfully assess robustness and adaptability~\cite{bender2021dangers}.



The difficulty of assessing AI models is compounded by the fact that some systems are not static; they continuously interact with dynamic environments and evolve based on new data and interactions. Evaluations conducted at a single point in time may fail to capture how these systems will perform under changing conditions or how they will adapt to new challenges~\cite{brahmakshatriya2023d2x}. This makes it difficult to design evaluations that account for the evolving nature of AI, as static tests cannot anticipate future states or emergent behaviors in the system~\cite{boyarskaya2020overcoming}. Furthermore, many AI systems operate in highly contextual environments where their performance can depend heavily on factors like user behavior, social norms, or regulatory provisions. In their work on evaluating interactive AI systems, \citet{boukhelifa2018evaluating} find that evaluating such models poses unique challenges: a system may respond to user preferences, but the users themselves alter their behavior based on system performance, leading to co-evolution effects that are difficult to untangle.

\subsection{History of Benchmark Use}
\label{sec:benchmark_hist}
To understand how benchmarks became a popular approach to AI system evaluation, we examine their history. Benchmarks originated as measures of computer hardware performance, intended to facilitate selection between an increasingly varied field of products \citep{lewis1985hist}. \citet{lewis1985hist} asserts that there is an important difference between benchmarking in the public and private sectors: the former is required to fairly and thoroughly compare potential vendors, thus placing a high importance on standardized benchmark assessments, while the latter prioritizes profitability and may favor cheaper and less rigorous approaches. As chronicled by \citep{church2018trends, liberman2010jelinek, raji2021benchmark}, both the public and private sectors drove the development of benchmarking for Natural Language Processing (NLP) tasks in the 1980s: DARPA funded the development of a well-defined, efficient evaluation for speech-to-text systems \citep{church2018trends}. This had a significant impact on the development of the field. As \citet{church2018trends} writes, a key benefit of this system was that it enabled hill climbing -- no longer limited to infrequent evaluations, researchers could measure and pursue more rapid advancements. IBM similarly worked to develop the Common Task Framework (CTF), which would become a precursor to modern AI benchmarking \citep{liberman2010jelinek}. By providing well-defined goals and corresponding tasks, the efforts of IBM and DARPA lowered barriers to entry, driving growth of a field \citep{liberman2010jelinek}. 

\subsection{Benchmarking in Other Fields}
We draw further insights about benchmark use from a review of benchmarking literature from other fields. 

\cite{cheng2022transistors} highlight the need for standardized benchmarks in the evaluation of field-effect transistors (FETs) to enable consistent and fair comparisons across different studies, given the complexity and variability of device structures and materials. According to the authors, the purpose of a benchmark is to provide a reliable framework for assessing key performance metrics, such as current, mobility, and contact resistance, to help guide the development and optimization of FET technologies. They note that benchmarks can also aid in decision-making by offering a clear basis for identifying the most promising technologies and guiding research efforts toward scalable and high-performance solutions.

\cite{chapman2018environmental} discusses the purpose and limitations of environmental quality benchmarks (EQBs). The primary function of EQBs is to serve as a tool to assess potential harm from chemicals and other stressors in aquatic ecosystems, though the author emphasizes they are not perfect and should not be used alone for final decision-making but that "they can provide understandable scientific input to decision makers". According to the author, the benchmarks are designed to filter out cases of negligible concern and highlight potential risks, aiding decision-makers in identifying priorities and guiding further investigations. 

Finally, \cite{aniba2010bioinformatics} discuss the role and purpose of benchmarks in bioinformatics, particularly focusing on their application in multiple sequence alignment. The primary function of a benchmark, as the authors state, is to provide an objective, comparative evaluation of computational tools by testing them against standardized datasets. This allows for the identification of strong and weak points, measuring improvements, and helping non-specialists choose appropriate tools. The authors also highlight that benchmarks in their field aid decision-making processes by enabling developers to refine their tools and helping users identify the best method for their needs, as benchmarks represent "a tool that is used to measure the ability of a software system to meet the requirements of the user".

\subsection{Theories of Information Technology Adoption}
\label{sec:theory_it_adoption}
Theories of information technology (IT) adoption study how new IT -- which broadly encompasses computers, software, data and information processing, and artificial intelligence -- becomes accepted and used. AI benchmarks themselves commonly combine software and datasets; as such, we consider them to be a part of IT and draw upon these theories to provide a framework for understanding when and how AI benchmarks are used. 

\citet{davis1986tam} introduces the \textbf{Technology Acceptance Model (TAM)}, which identifies two primary factors that impact the acceptance of information technology: perceived usefulness and perceived ease of use. The former refers to the perception of a given tool's ability to enhance an individual's job performance, while the latter refers to the perceived effort that the use of the tool requires. Integrating \citet{davis1986tam}'s model with several others, \citet{venkatesh2003unified} presents the \textbf{Unified Theory of Acceptance and Use of Technology (UTAUT)}. According to this framework, four constructs determine acceptance and usage: performance expectancy, effort expectancy, social influence, and facilitating conditions. Performance and effort correspond to \citet{davis1986tam}'s usefulness and ease of use, respectively. \citet{venkatesh2003unified} defines social influence as an individual's perception that important others believe they should adopt the technology in question. Finally, facilitating conditions are the degree to which an individual believes that organizational and technical infrastructure supports the technology's adoption. 

Applying this theory to AI benchmarks, we predict that \citet{venkatesh2003unified}'s four criteria must be met in order for benchmarks to be adopted. Since benchmarks aim to provide information about a model, we expect practitioners must determine this information to be useful in order for the criteria of performance expectancy to be met. The widespread use of benchmark scores in publications and marketing materials \citep{tascilar2023quest,bender2021dangers} suggests that social influence is fulfilled. However performance expectancy, effort expectancy, and facilitating conditions remain open questions that we explore in our interview study. We note that even when all factors are present, the UTAUT model does not provide a prediction about absolute levels of usage. Instead, it offers an interpretation for why benchmarks are or are not employed by practitioners.

\section{Methodology}
\label{sec:methodology}
\subsection{Research Design}
This study utilizes a grounded theory approach to investigate the development, use, and impact of AI benchmarks~\cite{anderson2005empirical}. Grounded theory is employed due to its suitability for exploratory research, enabling the derivation of theory based on empirical data~\cite{khan2014qualitative}. This method is particularly appropriate for examining dynamic and multifaceted subjects such as AI benchmarks, where established theoretical frameworks are limited or underdeveloped~\cite{heath2004developing}.
A semi-structured interview study was chosen for data collection. Interviews are a direct way to obtain evidence about attitudes, feelings, memories, perceptions, anticipations, goals, values, and the like, making them ideal for exploring participants' nuanced views on benchmarks~\cite{mcgrath1995methodology}. Semi-structured interviews provide an opportunity to gather rich, qualitative insights while maintaining the ability to adapt to the unique perspectives of each participant. This method provides flexibility for probing deeper into themes that emerged during the interviews, but were not necessarily anticipated by researchers~\cite{adams2015conducting}.

\subsection{Data Collection}
In-depth interviews were conducted with participants representing four categories of stakeholders: policy analysts, researchers in academia, researchers in industry, and benchmark users in product development and management. Interviews took place over the course of a month, September of 2024, via the online platform Zoom. The interviews were conducted in English and all interviewees were located in the United States.

Participant selection for this study followed a combination of intentional sampling and snowball recruitment. Initially, individuals were identified based on their recognized expertise and relevance to the subject matter, ensuring the inclusion of participants with diverse professional backgrounds. In addition, snowball sampling was employed, wherein participants were asked to recommend other potential informants who could offer insights. This was complemented by posting invitations on LinkedIn and X, which allowed reaching a broader audience of professionals. Finally, the recruitment process involved directly contacting experts through professional networks. Our work intentionally varies participants' backgrounds to allow for a broad perspective on the development, application, and impact of AI benchmarks across different domains~(see \cref{tab:demographics}). We note that our sampling approaches resulted in a population that was heavily skewed in terms of gender, thus limiting the generalizability of our conclusions. 

All participants provided informed consent, and their identities were anonymized in the final dataset to protect their privacy. IRB approval was obtained prior to data collection, ensuring the study adhered to relevant ethical guidelines for research with human subjects (IRB-76636).

\begin{table}[h!]
    \centering
    \small
    \begin{tabular}{>{\raggedright\arraybackslash}m{1.5cm} >{\raggedright\arraybackslash}m{1.5cm} > {\raggedright\arraybackslash}m{1.5cm} > {\raggedright\arraybackslash}m{4cm}} 
        \toprule
        \textbf{Informant} & \textbf{Gender} & \textbf{Area} & \textbf{Role} \\
        \midrule       
        I-1& Male & Policy & Research Scientist\\
        I-2& Male & Policy & Research Scientist \& PhD Student\\
        I-3& Female & Policy & Senior Manager \\
        I-4&  Female & Product & Nurse\\
        I-5& Male & Product & Chief Technology Officer\\
        I-6& Male & Product & Machine Learning Manager\\
        I-7& Male & Product & Software Developer\\
        I-8& Male & Product & Research Scientist\\
        I-9& Male & Product & AI Engineer\\
        I-10& Male & Product & Senior Engineer\\
        I-11& Male & Research & PhD Student\\
        I-12& Male & Research & Associate Professor\\
        I-13& Male & Research & PhD Student\\
        I-14& Not Specified & Research & PhD Student\\
        I-15& Male & Research & PhD Student\\
        I-16& Male & Research & Professor\\
        I-17& Male & Research & AI Team Lead\\
        I-18& Male & Research & Research Engineer\\
        I-19& Male & Product & Venture Capitalist\\
        \bottomrule
    \end{tabular}
    \caption{Informants}
    \label{tab:demographics}
\end{table}

\subsection{Interview Design}
The interview design for this study employed a semi-structured format, allowing for both consistency in data collection and flexibility for exploring emerging themes in greater depth. Interview questions were crafted to align with the study’s objectives of examining the development, application, and impact of AI benchmarks across different domains. This approach enabled the participants to share their experiences while allowing the interviewer to follow up on specific insights as they came up. The interview guide (see \cref{app:interviewguide}) was structured into multiple key sections, with questions designed to gather comprehensive data on participants' interactions with AI benchmarks and their perspectives on their effectiveness, limitations, and improvement suggestions.

We began our interviews with introductory questions such as, "Please describe your current role" and "How do you interact with AI models in this role?" These questions were included to establish a baseline understanding of the participant’s professional background and their direct engagement with AI systems. By gaining insight into each participant's unique context, the interviewer could better interpret the interviewee's responses to subsequent questions about benchmarks.

Next, the interviews moved to questions like "In the context of your work, what is the purpose of AI benchmarks?" and "How satisfied are you with the ability of existing benchmarks to fulfill this purpose?" which were designed to probe participants’ views on the functionality and relevance of benchmarks in their professional tasks. These questions were critical for assessing the alignment between current benchmarks and their intended goals. Additionally, questions such as "Can you describe your process for finding a suitable benchmark?" and "What factors are you considering when choosing a benchmark?" aimed to capture the decision-making process involved in benchmark selection, highlighting the criteria and challenges faced by practitioners.

Further questions, such as "What risks and benefits do you associate with using benchmarks for decision-making?" were designed to understand the broader implications of benchmarking practices, including their impact on outcomes and the potential risks associated with relying on benchmarks. The inclusion of these questions provided a balanced view of the perceived benefits and limitations of benchmarks, offering insights into potential areas for improvements. The structure of the interview guide allowed for a comprehensive exploration of AI benchmarking, while maintaining flexibility to explore unanticipated but relevant themes that arose during the conversations.

Before formal data collection, the interview protocol underwent pilot testing with three participants. Post-interview team debriefs identified which questions worked well, which were unclear, and which elicited unexpected but valuable themes. To achieve this, the team engaged in a structured feedback analysis process~\cite{mcgrath1995methodology}. Interview questions were assessed for clarity, relevance, participant engagement, and depth of response. Team debriefs and revisions were logged for transparency. Based on this feedback, we refined the interview guide, rewording and reordering questions to maximize clarity and relevance. All interviews were recorded via Zoom and transcribed using Otter.ai.

\subsection{Data Analysis}
Data was analyzed using grounded theory coding. Interviews were scheduled for 45 to 60 minutes and transcribed, resulting in a total of 517 pages and 141,702 words, with the per-interview word count ranging from 3,932 to 10,926. During the analysis, we extracted 521 passages (primarily individual sentences) and the process began with open coding~\cite{corbin2015basics,muller2014curiosity}, where key themes and concepts were identified from the raw interview transcripts. 

As the analysis progressed, axial coding was employed to explore the relationships between these themes, leading to more cohesive categories. The final set of 13 high-level code categories and 85 codes including descriptions and the number of corresponding tags in the interviews can be found in Appendix \ref{app:codinghandbook}.
The analysis followed an iterative process, with the team continually revisiting and revising the codes based on emerging data. A thematic analysis approach ensured that the study captured the richness of the interview data while building a grounded theory that reflects the participants' experiences. Multiple coders were involved in the process to ensure rigor and reliability, with coding discrepancies resolved through regular team discussions.
We used Otter.ai for transcription and Taguette for qualitative coding and thematic analysis. These tools facilitated efficient handling of the data while allowing for transparent tracking of revisions and changes during the coding process.

\section{Findings}
\label{sec:findings}
Through qualitative analysis of our interview data, we found that although all participants used benchmarks for relative comparisons of models, few interpreted these scores as an absolute signal, relying on supplemental measures to make downstream decisions. In this section, we present our detailed findings, highlighting how the strengths and weaknesses of existing benchmarks lead to the observed use patterns.

\subsection{Relative Measures to Assess Progress}
\label{sec:measure_relative}
Across all three settings considered -- research, product, and policy -- participants reported that benchmarks were highly useful as standardized measures of relative progress towards a benchmark's target metric. 

I-11 describes the ability of benchmarks to evaluate whether a new method is an improvement over previous ones:
\begin{quote}
    "So to make progress, when you're designing a method . . .  you need to have a sense of if you're actually heading in the right direction. And I think benchmarks are often nice if they actually do encapsulate the problem you care about."
\end{quote}

The relative assessments benchmarks provide can guide future development by indicating the efficacy of current approaches. I-15 describes benchmarks as a directional measure: "We're using them as sort of a vector, right? Are we headed in the right direction?" This type of insight is also beneficial in a product setting. I-6 lists questions benchmarks can answer about a new model: 
\begin{quote}
    "How do we do [with a] previous version of a model compared to a new version, or . . . if you're doing a feature improvement, how does the addition of a feature improve or decrease the performance?"
\end{quote}
Practically, these signals guided model developers and users in decisions such as whether to adapt the model (e.g., if benchmark scores dropped precipitously) or to continue development (e.g., if it outperformed some baseline). These relative scores were more often used as negative indicators than positive ones. In other words, poor benchmark performance was more likely to disqualify a model from further development or use than good benchmark performance was to qualify it. I-8 describes the impact of benchmarks on model development:
\begin{quote}
    ""We'll see regressions [in score] and then that's led to some debugging. For instance, people say 'why did numbers get worse on this dimension?' So they go back, 'Oh, actually, we made a mistake' . . . Or, actually, you know, we had tried a more aggressive approach for this thing . . . and we need to roll things back. . . . Yeah, it's called some late nights for someone on our team too, where they find out, like, . . . oh, actually, the results weren't as good as we expected, so we need to try to slip in another checkpoint before things get released.
\end{quote}
In some cases, negative performance could permanently halt a model's development. I-18 called the impact of significantly poor benchmark performance a "hammer blow," since it could effectively terminate a project. 

Across all three settings, participants agreed that the standardized, relative measures benchmarks provided were useful as an initial indicator of model performance. They also aligned in considering a model to have failed if its performance significantly dropped in comparison to some baseline. However, when benchmark scores improved, research benchmark users differed from those in product and policy in how they answered the question: is better good enough? 


\subsubsection{Research: Is Our Model Better?}
\label{sec:research_better}

When used in research settings, benchmarks measured comparative performance between old and new methods. I-18 underscores the importance of the standardized measurement these benchmarks provide: 
\begin{quote}
    "Without it, you won't even have any kind of feedback loop, right? So we won't know how we are doing against . . . the state of the art models, even if we come up with a new technique."
\end{quote}

These evaluations were accepted to the extent that I-17 described benchmark performance as essential to publication: "If you're trying to publish an academic paper and it doesn't beat the benchmark, it won't get published." Within academia, these scores could be viewed as definitive signs of function and progress, as reflected in I-15's statement that "the goal in academia is to just push the number higher." In research, product, and policy, benchmarks were considered markers of relative improvement. A model was good if it outperformed baselines on benchmark assessments. This was a key point that distinguished research from product and policy: in research, improvement on a benchmark alone made a model good.   

\subsubsection{Product and Policy: Is Better Good Enough?}
\label{sec:prod_pol_enough}
Confronted with the consequences of real-world deployments, product and policy practitioners in our interviews required a more rigorous standard. I-1 explained this need:
\begin{quote}
    "If you want to build benchmarks that inform policy or inform a societal response to AI systems, that requires having an . . . understanding of the absolute capabilities, rather than the relative capabilities."
\end{quote}
Given the lack of absolute signal available, I-7 observed that benchmarks were not used as the basis of significant decisions: "People look at them. They make minor code changes based on them. I've never seen someone launch or not launch, like, or do anything super critical based on a benchmark." When industry benchmarks were used to inform deployment, this was cause for concern: "I get very nervous when, when people say . . . 'based on this benchmark, we think this is launchable'" (I-8). The most notable gap, in I-8's perspective, arose because "it's impossible to know how people are going to use the [model being evaluated]". To mitigate this, he suggested that model developers draw from the release processes used for other products: "you have test users come in, or you do . . . user case studies." This need for substantial additional assessment contrasts starkly with the research setting, where benchmark improvement alone was considered a success. 

\subsection{General Assessments, Specific Applications: When Benchmarks Lack Relevance}
\label{sec:general_specific}
The observed discrepancy in benchmark interpretation may arise from the fact that unlike decisions between models made in research settings, those made in policy and product often require absolute signal due to their potential for direct impact on people. Many of the participants we spoke to found publicly available benchmarks lacking and were actively developing benchmarks suited to the specific deployments they were concerned with. Based on our analyses, we identify the gap between what existing public benchmarks assess and how models will be practically used as a key factor that limits their meaningful adoption. I-9, who was building benchmarks tailored to specific industrial applications, cited the fact "that benchmarks today are not based on the end user behaviors which are most interesting or especially the most valuable" as a "key deficit." Assessing the examples within popular public benchmarks, I-7 found them to be not only irrelevant, but also stated
\begin{quote}
    "Some of the major benchmarks that people use today, MMLU, for example, . . . just go through the questions and look at them, and you'll realize that they are awful. Just like, . . . terrible. Totally, a bunch of totally irrelevant stuff, things that are totally wrong. "
\end{quote}


While some participants developed their own benchmarks to address issues of quality and relevancy, others forewent benchmarks entirely. Although I-5 noted there were no existing "benchmarks for the kinds of data that . . . end customers care about," this did not lead his company to develop its own benchmarks. Due to the level of detail required to understand their models' performance, the company decided not to "do anything programmatically today, because oftentimes, [to] improve, the changes can be . . . somewhat nuanced." Manual evaluation of these specific qualities was seen as more useful than automated evaluation of general ones.

\subsubsection{The specific case of safety-critical systems}
\label{sec:safety_critical}
Such evaluation was not considered sufficient by all practitioners. When issues of safety were at stake, policy and product workers needed assessments that were both standardized and grounded in evidence and downstream use cases. A goal of I-8, who developed benchmarks for safety-relevant features, was "trying to anticipate the real user behavior." I-1 described the challenge of evaluating risks that arise from a model's downstream use: 
\begin{quote}
    "Proxies are like doable, and . . . relative comparisons are doable, but absolute sort of assessment of a model's risk is a lot harder because . . . there's a difference between the things you can do in a benchmark and how a model will actually be used."
\end{quote}
General public benchmarks were not considered suitable for such applications. I-14, who was researching benchmarks to evaluate AI models for medical applications, described the gap between performance on such benchmarks and performance on downstream tasks: 
\begin{quote}
    "Large language models are benchmarked on some multiple choice questions, but we found that those don't actually reflect well when applied to clinical settings, because .. . the proxy that was used to create the benchmark as a multiple choice question was not . . . reflective of how things function in the clinic"
\end{quote}
In I-8's case, since safety was defined in the context of company policy, public benchmarks were not suitable because their "policy is specific to our company." 

For example, R4 said that "[our team] developed our own benchmark, because we felt that people weren't talking about a very important aspect of medicine." When safety was a concern, it was necessary to develop benchmarks that were not only specific to some task, but also grounded in the potential consequences of that task's automation.

\subsubsection{Informative Benchmarks Fit Real-World Use Cases}
\label{sec:informative_real_world}
Outside of safety-critical use cases, the most informative benchmarks were also considered to be those where the ability being measured was tied as closely as possible to the real-world use case of the model being evaluated. Without such grounding, benchmark scores were difficult to interpret. I-10 compared assessing models' ability based on a single quantitative benchmark score to assessing people's health based on weight: 
\begin{quote}
    "I view it as like . . . measuring your body weight. It's a great measurement week to week, but your body weight and mine are not comparable, right? Like you may have a different BMI. I may have different health issues. So that's kind of a thing where . . . we try too hard to generalize it when it's not generalizable."
\end{quote}

I-11 similarly noted that there may be difficulty translating general benchmark scores to information about applied use: "benchmarks are probably very predictive and . . . can give you signal of . . . how progress is going, but there's always . . . a benchmark to real world gap." To address this gap, I-7 emphasized the importance of designing benchmarks with particular use cases in mind: "if you want to understand which model is going to work best for a particular task, you basically just need to test it for that task." 

Benchmark developers found that incorporating both model users and domain experts in the design process was critical to creating benchmarks that were truly suited to real-world use. On incorporating users in this process, I-14 said
\begin{quote}
    "It's very important if you're looking on something that is relevant to a community that you know little about, or you may not exactly be very immersed in. You probably don't have all the feedback that's necessary to build that internally. You will know all the things, so engaging with community . . . is very, very important, not just to pick the task, but also to pick the metrics that define that task."
\end{quote}
One way this feedback could be incorporated into benchmark design, is by having "some set of human validation on how closely that metric is mimicking what you would really want" (I-4). Like I-14, I-1 emphasized the importance of consulting individuals in the relevant domain to account for knowledge which benchmark developers may lack. He described the importance of working with domain experts:
\begin{quote}
    "You have to be an expert to know an expert. And so a lot of the capabilities you're interested in are highly specific, and it's very hard to build evaluations of a task that you can't do yourself. Having experts that can do the task, can reason about the task, can reason about right or wrong answers by models, is crucial in developing useful benchmarks . . . otherwise, you're sort of walking around in the dark"
\end{quote}
I-1 similarly incorporated feedback from these experts into the design of metrics. He outlined the process followed: 
\begin{quote}
    "I gave infrastructure support to these experts. Part of the question design process was designing rubrics with which to grade the questions. And I helped with that and created a loop where they could change the rubrics and see what the effect was on the grading system"
\end{quote}
Both I-1 and I-14 directly included stakeholders in the metric development process, ensuring that the traits their benchmarks measured were correlated with domain experts' judgments. This not only grounded the benchmarks in model use cases, but also filled gaps in developers' knowledge.

Valid measurement of the model's desired traits emerged as a key factor in adoption. The participants who found benchmarks to be most effective were those who were most directly able to measure the abilities of interest. R8, who developed automatic speech recognition (ASR) models, stated that for "self supervised pre-training, the only way to know if that technique is working is by . . . comparing it on ASR benchmarks." These benchmarks were able to be helpful since the metric they used, word error rate, was well-aligned with the goal of recognizing speech. 

\subsection{Benchmarks Cannot Replace Human Evaluation}
Even when satisfied with available automated metrics, participants in all three groups agreed that benchmarks should be complemented by, rather than substituted for, human evaluation. I-7, who found benchmarks to be highly suited to his use case of ASR, said 
\begin{quote}
    "That would be the ideal scenario where you get someone speaking or each language, which we cover, and asking them to spend, like, a couple of hours giving them some instructions where they just absolutely go ballistic at the model, do all kinds of things and see that you will get, like, a good trend."
\end{quote}
When assessing the medical models they were developing, I-14 found that "keeping humans in the loop is necessary." Human feedback could supplement both benchmarks and approaches such as model-based red-teaming. A robust system, I-8 reported, would "have . . . red teaming, and also just . . . people [to] test out the models, like, use it in whatever way, as additions to benchmarks." Participants agreed that there were insights which could only be gained through direct, manual evaluation of the models. 

An additional benefit of such human evaluation is that it is durable in the face of rapid model advancement. While benchmarks frequently saturate, human evaluation remains relevant. I-2, whose work concerned the assessment of future model capabilities, said 
\begin{quote}
    "I place a high value on, like, the human interaction, vibes-based evaluation technique, because it's. . . more or less the only thing that doesn't get invalidated by a new paradigm or type of model. We don't really know what models will be like in like five years or something, and it would be very surprising if many of our benchmarks were still used then."
\end{quote}
I-11 named human-centered evaluation as the most significant gap in existing benchmarks: 
\begin{quote}
    "Benchmarks. . . measuring like, how LLMs or just generally, models interact with humans are missing. It's hard and there's not really a clear setup. I think people probably need to partner more with . . . psychologists and social scientists on how to like do these benchmarks. But those would be the most valuable." 
\end{quote}

Participants agreed that task-specific evaluations, designed with a model's use case in mind, were more helpful than generic ones. Consistent with this belief, they expressed that for models directly used by humans, human evaluation is irreplaceable. Although some participants validated their benchmark results with extensive human evaluation, such testing was most commonly performed by the developers themselves, who interacted with their own models during the development process. Many noted the expense of human annotation as a significant deterrent for collecting comprehensive manual evaluations.

\section{Discussion}
\label{sec:discussion}



\subsection{AI Benchmark Adoption Through the Lens of UTAUT: Performance, Effort, Social Influence, and Facilitation}

\label{sec:utaut}
Our work seeks to understand how and when AI benchmarks are used by practitioners. Deciding to use a benchmark can be considered the acceptance of a technology; thus, we return to the UTAUT framework and discuss which of its criteria were or were not met for our participants. Applying this framework, we see that although public benchmarks were generally considered easy to use (effort expectancy), were widely accepted (social influence), and generally came with infrastructure to support their use (facilitating conditions), individuals working in policy and product did not believe them to provide sufficient information to be helpful (performance expectancy). In the following section, we examine factors which contribute to this low performance expectancy and propose changes to the benchmark design and development process that can increase relevancy and improve performance as real-world measures, thereby enhancing both adoption and quality.

More general models have been matched by broader benchmarks testing general, increasingly abstract model capabilities; however, such evaluations face significant issues of construct validity and real-world applicability \cite{raji2021benchmark, cronbach1955construct}. Assuming that construct validity in general evaluations was no longer a problem, the question remains: are general benchmarks the most useful approach? When considering a candidate for a specific job, their experience with and approaches to its intended tasks are more important than broad traits such as intelligence. Similarly, if models are to be deployed for particular applications, then it is these use cases which most require evaluation, rather than more general capabilities. As discussed in \cref{sec:general_specific}, our participants desired and often worked to create benchmarks that were more task-specific or real-world use-case inspired, since they found broader benchmarks did not meet their needs.

\cref{sec:benchmark_hist} describes how benchmarks as we know them today originated from a need to make standardized comparisons between vendors considered for government contracts \citep{lewis1985hist, raji2021benchmark}. Although hardware contexts still present challenges of ecological validity \citep{lewis1985hist}, evaluation of AI systems poses even greater difficulties. 
Hence, in the absence of application-oriented designs, benchmarks can only provide relative signal. Across research, product, and policy, this was our participants' most frequent use case for AI benchmarks. They were confident that benchmarks could demonstrate improvement, but almost always required information and assessment beyond benchmarks when assessing absolute capabilities or properties to make substantive decisions. This did not make benchmarks useless -- as I-15, an academmic researcher said, "It would be impossible to do my job . . . without them." He explained this necessity in terms of standardization: with benchmarks, "the whole community knows you measure the same thing." Furthermore, many participants identified benchmarks as a key factor in driving AI advancements. "The field [of AI]," I-15 stated, "has progressed because our ability to make benchmarks." The mere existence of a benchmark could cause a subfield to advance, since, as I-14 stated, "Having a public data set . . . attracts technical people to work on that problem." 

Despite their power to drive fields forward, participants remained hesitant to use them for substantive decisions and identified overreliance on benchmark scores for decision-making as a substantial risk. I-9 outlined this danger:
\begin{quote}
    "The . . . most obvious risk is that there's a disconnect between [a] domain level benchmark and a task specific benchmark. What you see at a very general level is a trend that doesn't apply to your actual use case. Another issue that we see a lot of is kind of like understand[ing] interpreting the results effectively. So you have a benchmark of legal performance, and the models are all performing in some range of like, 70 to 80\% does that mean that they are doing very well? We should rely on them, or should we wait until they get closer to 95, 99\% before they're acceptable for use. Additionally, like, how does it compare to human performance? If humans are doing like 60\% then . . . they're actually doing very well"
\end{quote}
Under the UTAUT, this quote reflects the principle of performance expectancy. Participants were only comfortable adopting benchmarks under narrow circumstances, reflecting their doubts about the insights these assessments could provide. 

While benchmarks lacked performance expectancy, their acceptance was heavily supported by social influence. \citet{venkatesh2003unified} defines this factor as "the degree to which an individual perceives that important others believe she should use the new system." Benchmark use as an AI community norm is reflected in both the necessity of using benchmarks for research publication and the reporting of these scores in model marketing materials \citet{openai2023gpt4, anthropic2024claude3}. I-13 noted that "if you're evaluating LM capabilities, you're going to go use like MMLU and MT bench and AGI Eval, because everyone uses them" but not because they hold inherent value. Social influence may compel model developers to run benchmarks, but it does not necessarily drive model users to rely on these scores.  I-19, a venture capitalist investing in AI companies, noted that in addition to being a poor signal for the success of an early stage company, benchmarks were unhelpful since "public benchmarking has become like propaganda." The fact that participants generally did not use benchmarks to make their ultimate decisions and highlighted the need for additional information contrasts with the way benchmark scores are often used in marketing materials. This finding is consistent with the UTAUT model -- according to \citet{venkatesh2003unified}, performance expectancy is the most important criteria for technology adoption, while social influence has more complex and less consistent effects. Benchmarks may have been accepted as a preliminary measure; however, nearly all participants in product and policy supplemented them and some dismissed them entirely. 

\subsection{Improving Benchmarks is Necessary -- and Hard}
Despite benchmarks' widespread use and well-known limitations, participants who developed benchmarks in both research and product settings found that the impact and difficulty of their work was underestimated. I-8 described his colleagues' perception of benchmarking: "I don't think it's appreciated enough. I think people say, like, just come up with the data set. How hard can it be?" In research, I-15 expressed that such attitudes led some students away from benchmark development:
\begin{quote}
    "My biggest gripe . . . is going to be that our field only exists and progresses because of benchmarks. Like, if you don't publish experimental results on a benchmark, you don't publish like, you literally don't publish. You will get rejected. Yet, if you try to publish a benchmark data set at a conference, it'll more likely than not be rejected because you didn't present a method . . . So if you . . . propose that let's build a new benchmark, everyone will be like, well, let's hold on a second. Maybe that's not worth the time . . . There's no ROI on that for a PhD student."
\end{quote}

Understanding the difficulty of developing benchmarks is critical, since the problems with existing methods cannot be solved without the investment of substantial time and resources. Practices like the inclusion of community members and domain experts in the benchmark design process are costly, but are essential to evaluating model performance in a way that is truly meaningful to those that models most directly impact. 

\subsection{What Is A Better Benchmark?}
\label{sec:better_bench}

Many of the participants in our study were working to mitigate the outlined issues with existing benchmarks by developing new, internal ones. When asked about how they addressed these challenges, participants emphasized that a good benchmark should offer a meaningful evaluation of an AI system’s capabilities, aligning closely with real-world applications to ensure robust, meaningful assessments. I-2 mentioned that benchmarks "should be actually informative about [a] particular capability" and I-7 stressed the importance of "making sure the data is realistic." I-7 emphasized that "there's a ton of little domain-specific things that matter a lot to getting a good benchmark," highlighting the need for benchmarks to address the specific contexts in which AI models will be deployed, and the necessity of involving domain experts in the design process. 

Another trait that participants emphasized was that of transparency in benchmark scope and goals as a step towards preventing misinterpretation. One way to make scores easier to interpret, which participants frequently mentioned, was providing a human baseline for comparison. For instance, I-2 highlighted that "assessing the model's capability relative to the amount of time it would take a human to complete a similar task" was useful. Without this context, the interpretation of a benchmark result was seen  not only as difficult, but as a potential risk when unclear results were used for downstream decisions.

The consensus was that benchmarks need to be carefully designed to match the distribution of real-world data and produce statistically significant, reproducible results. Diversity, data quality, and an appropriate level of difficulty were viewed as important factors in ensuring that benchmarks remain informative and adaptable to evolving AI capabilities, avoiding the trap of saturating too quickly: 

\begin{quote}
    "Having a benchmark that you can get signal on today, where models capabilities vary in a meaningful way, but that won't saturate very soon, either is, I think, quite useful, and we've seen with a lot of benchmarks, that they saturate quite quickly. So this just means having a wide range of difficulties for your tasks included in the benchmark." (I-1)
\end{quote}

Moreover, participants acknowledged several ongoing challenges in the design of effective benchmarks. One key issue is the difficulty in defining clear, interpretable metrics, and I-8 noted about their internal evaluations that 

\begin{quote}
"We also rarely just report one number, because a lot of this is trade-offs, like one number goes up while the other number goes down, for instance. So we shy away from that. It drives a lot of the developers nuts, right? They just want a single number to optimize, but then there's Goodhart's law. [...] Okay, what's the final number? Like, there's no final number." 
\end{quote}

Another frequently voiced concern was the contamination of datasets and the need for proprietary data collection methods to mitigate this risk, which was mentioned by participants as a key reason for not releasing internally developed benchmarks: 

\begin{quote}
"[We are] actually trying to make sure that when we collect data sets, we keep them proprietary, we don't release them publicly, and we make sure that all of the model providers we use will not retain or retrain on any of the data." (I-9)
\end{quote}

These findings are in line with previous work on AI benchmark quality: \cite{reuel2024betterbench} have shown that many public benchmarks do not take preemptive measures against contamination, outline how benchmark results should be interpreted, or thoroughly explain how knowing about the tested concept is helpful in the real world. Similarly, \cite{liu2024ecbd} find that existing benchmarks often lack validity, which impacts their real-world usefulness. 

One overarching challenge respondents emphasized was that the science behind robust evaluations and benchmarks is "an open problem. How to figure out, like figuring out what's a reasonable benchmark?" (I-8). Our findings suggest that deliberate design processes, including the improvement points stressed by participants, are necessary to ensure that benchmarks provide meaningful evaluations to practitioners. 

Ultimately, benchmarks must move beyond being a "marketing strategy" to serve as robust tools to practitioners for evaluating AI systems in a transparent, informative, and interpretable manner that reflect meaningful improvements in model capabilities rather than artificial performance gains.

\section{Limitations}
\label{sec:limitations}
The study's scope is limited by its reliance on self-reporting methods, such as interviews, which may introduce bias due to participants' desire to appear competent or answer in socially desirable ways~\cite{nederhof1985methods}. The presence of the researcher and the potential for the researcher’s perspective to influence the interview process are further acknowledged as potential sources of bias~\cite{lofland2022analyzing}. In addition, the study might suffer from selection bias, resulting in a selection of interviewees that is not reflective of the underlying population of AI benchmark users. In particular, while we specifically included people who indicated that they consciously decided against using AI benchmarks in their work, these respondents made up a significantly smaller fraction of our interviewees. We also acknowledge the gender bias in our sample; while we tried to have a more balanced interviewee selection, our chosen sampling techniques favored a larger male interviewee proportion, given that research and development of AI systems are overall male-dominated fields. This could have been mitigated through a stratified sampling approach and should be considered when drawing conclusions from this work, as they may not generalize to a more balanced population. Future work should consider incorporating additional data collection methods, such as surveys or observational studies, to complement the interview data.

\section{Conclusion}
AI benchmarks can act as markers of relative progress, especially in academia, but fail to allow new models to demonstrate actual increased performance on relevant capabilities. 
Marketing materials may assign substantial meaning to benchmark scores; however, most benchmarks provide limited information regarding practical deployment decisions. Although a score of 80 is relatively better than a score of 70, figures such as these are only meaningful in the context of a benchmark designed to inform downstream use. Given the high-stakes settings in which AI models are increasingly used, it is critical that practitioners have informative comparative measures available to them. Through measures such as the incorporation of community members and domain experts in the benchmark design process, more relevant benchmarks can be developed. Based on the interviews we conducted, we propose that the difficulty of creating such benchmarks must be appreciated so that resources can be allocated appropriately and that even ideal benchmarks cannot obviate the need for human evaluation. 
\label{sec:conclusion}

\bibliographystyle{ACM-Reference-Format}
\bibliography{iui}

\newpage

\appendix
\section{Appendix}
\subsection{Interview Code Data}\label{app:codinghandbook}

\begin{longtable}{|l|p{7cm}|c|}
\caption{Interview Code Data} \\
\hline
\textbf{Title} & \textbf{Description} & \textbf{Count} \\
\hline
\endfirsthead

\hline
\textbf{Title} & \textbf{Description} & \textbf{Count} \\
\hline
\endhead

\hline \multicolumn{3}{|r|}{\textit{Continued on next page}} \\
\hline
\endfoot

\hline
\endlastfoot

01.use.decisionmaking & How benchmarks influence decisions & 4 \\
\hline
01.use.decisionmaking.finetuning & Additional training or finetuning based on benchmark results & 14 \\
\hline
01.use.decisionmaking.selection & Model selection based on benchmark results & 7 \\
\hline
01.use.frequency & How often do they use a benchmark? & 1 \\
\hline
01.use.frequency.daily & Use benchmarks daily & 3 \\
\hline
01.use.frequency.infrequently & Use benchmarks infrequently & 0 \\
\hline
01.use.frequency.monthly & Use benchmarks monthly & 1 \\
\hline
01.use.frequency.other & Frequency of benchmark use not captured by other codes (unique) & 0 \\
\hline
01.use.frequency.weekly & Use benchmarks weekly & 3 \\
\hline
01.use.impact & How benchmarks affect outcomes & 2 \\
\hline
01.use.impact.direct & Decision directly made based on benchmark result & 10 \\
\hline
01.use.impact.influence & Benchmark result influenced decision & 14 \\
\hline
01.use.impact.none & Benchmark result did not influence decision & 2 \\
\hline
01.use.information & Information they are seeking from the benchmark & 7 \\
\hline
01.use.information.customer & No information but request from customer to use benchmark & 4 \\
\hline
01.use.information.informtraining & Identification of additional data needed & 14 \\
\hline
01.use.information.popular & No specific information needed but done because it's popular & 0 \\
\hline
01.use.information.usecase & Identification of best model for use case & 6 \\
\hline
01.use.information.validation & Validation of developer information & 2 \\
\hline
01.use.information.weakness & Identification of weaknesses of a model & 16 \\
\hline
01.use.purpose & What is the benchmark used for? & 14 \\
\hline
01.use.trends & Changes in benchmark use over time & 7 \\
\hline
01.use.trends.change & Change over time observed & 12 \\
\hline
01.use.trends.nochange & No change over time observed & 0 \\
\hline
01.use.information.capabilities & Identification of capabilities of a model & 19 \\
\hline
01.use.methodology & How they are using a benchmark (methodologically), e.g., do they run it multiple times? & 5 \\
\hline
01.use.information.threat & Identification of a subsequent threat through a model/model use & 8 \\
\hline
01.use.information.humantime & Additional benchmark metric that tracks how much faster a model can do a task compared to a human & 2 \\
\hline
01.use.information.cost & Additional benchmark metric that tracks how much completing a task costs for a model & 1 \\
\hline
01.use.information.monitor & Use benchmarks to monitor performance of a model while it is deployed. & 6 \\
\hline
01.use.impact.others & Use benchmarks because informs others about their model & 3 \\
\hline
01.use.information.policy & Use benchmarks to inform policy & 7 \\
\hline
01.use.information.other & Seeking information from benchmark not captured in set of tags & 10 \\
\hline
01.use.information.compare & Use benchmarks to compare the performance of two models in a standardized way & 18 \\
\hline
01.Use.information.progress & Benchmarks inform model improvement over time & 9 \\
\hline
01.use.ideal & Mentions how benchmarks should be used & 4 \\
\hline
01.use.decisionmaking.deployment & Decision to use model for a use case or not based on benchmark results & 3 \\
02.selection.criteria & Factors considered when choosing a benchmark & 5 \\
\hline
02.selection.criteria.customer & Chosen based on customer need & 2 \\
\hline
02.selection.criteria.familiar & Chosen because it was used before & 3 \\
\hline
02.selection.criteria.popular & Chosen because it's widely used & 6 \\
\hline
02.selection.criteria.usecase & Chosen for specific use case relevance & 9 \\
\hline
02.selection.recent & Details about a recently chosen benchmark & 1 \\
\hline
02.selection.sources & Where information about benchmarks is found & 6 \\
\hline
02.selection.specific & Mentions of specific, publicly available benchmarks & 29 \\
02.selection.criteria.opensource & Chosen because it was opensource & 2 \\
\hline
02.selection.criteria.multiplechoice & Chosen because it was multiple choice format & 1 \\
\hline
02.selection.criteria.design & Chosen because it was well designed & 1 \\
\hline
02.selection.criteria.trustdeveloper & Chose benchmark because trusted the benchmark developer & 3 \\
03.types.bias & Benchmarks used to assess bias in AI models & 2 \\
\hline
03.types.performance & Benchmarks that measure model performance & 14 \\
\hline
03.types.safety & Benchmarks used to evaluate safety aspects of AI models & 20 \\
\hline
04.satisfaction.context & Whether benchmarks provide enough context for interpretation & 17 \\
\hline
04.satisfaction.level & Expressions of satisfaction or dissatisfaction with existing benchmarks & 21 \\
\hline
04.satisfaction.usefulness & Assessment of how useful benchmarks are & 20 \\
\hline
05.gaps.improvement & Suggestions on how existing benchmarks could be improved & 32 \\
\hline
05.gaps.lacking & Areas where more benchmarks are needed & 35 \\
\hline
05.gaps.sufficient & Areas where there are enough benchmarks & 14 \\
\hline
05.gaps.issues & Areas where there are issues with existing benchmarks & 76 \\
\hline
05.gaps.relevant & No relevant benchmarks available, even if there exist benchmarks for a use case & 8 \\
\hline
05.gaps.modality & Lack of benchmarks for a certain modality (audio, vision, etc.) & 3 \\
\hline
05.gaps.deterrent & Deterrent from using or developing a benchmark & 15 \\
\hline
05.gaps.policy & Gaps in regulatory provisions that guide the use of benchmarks & 13 \\
\hline
05.gaps.opensource.v.pro & Observed gaps between the quality of benchmarks created opensource vs. larger professional labs & 3 \\
\hline
05.gaps.issues.use & Issues in how benchmarks are being used & 6 \\
\hline
06.risks.rewards.mitigation & Strategies to mitigate risks associated with benchmarks & 18 \\
\hline
06.risks.rewards.rewards & Benefits or positive outcomes of using benchmarks & 8 \\
\hline
06.risks.rewards.risks & Mentioned risks or downsides of using benchmarks & 36 \\
\hline
07.development.collaboration & Mentions of collaborative benchmark development efforts & 19 \\
\hline
07.development.ideal & Descriptions of what an ideal benchmark would look like & 47 \\
\hline
07.development.design & Mentions design considerations for designing a good benchmark & 48 \\
\hline
07.development.gaps & Description of what is missing in the benchmark design process & 6 \\
\hline
07.development.challenges & Description of the challenges for benchmark developers & 4 \\
08.application.realitygap & Discrepancies between benchmark and real-world performance & 18 \\
\hline
08.application.tradeoffs & Tradeoffs between general and use-case specific benchmarks & 3 \\
\hline
09.reflection.field & Trends in benchmark use in the field more broadly & 47 \\
\hline
09.reflection.historical & How benchmark use has changed over time for individual or organization & 2 \\
\hline
09.reflection.field.incentives & How the incentives behind using benchmarks have evolved over time & 3 \\
\hline
10.category.external & Externally sourced or industry-standard benchmarks & 34 \\
\hline
10.category.internal & Internally developed or used benchmarks & 33 \\
\hline
10.category.internal.to.external & Developed benchmark internally but it is now opensource, and they continue to use it. & 4 \\
\hline
10.category.product & Developed benchmark for a customer as a 3rd party AI evaluator & 3 \\
\hline
11.benchmark.def & Defines benchmarks theoretically & 2 \\
\hline
12.metric.ideal & What characteristics make an ideal/good metric & 3 \\
\hline
zz\_intro & Background on interviewee's current role & 28 \\
\hline
\end{longtable}

\subsection{Interview Question Guide}\label{app:interviewguide}

\textbf{Introduction and Background}
\begin{enumerate}
    \item Please describe your current role.
    \begin{itemize}
        \item How do you interact with AI models in this role?
        \item How did you come to this role?
        \item Can you tell me about your relationship to AI and machine learning technology more broadly?
        \item How does your educational background inform the way that you think about AI?
    \end{itemize}
\end{enumerate}

\textbf{Benchmark Usage and Purpose}
\begin{enumerate}
    \item In the context of your work, what is the purpose of AI benchmarks?
    \begin{itemize}
        \item How satisfied are you with the ability of existing benchmarks to fulfill this purpose?
    \end{itemize}
    \item What are you using benchmarks in your job for?
    \begin{itemize}
        \item Can you describe your process for doing this?
        \item Of the use cases you mentioned, which is the most prevalent?
        \item What is your goal in using benchmarks for these purposes?
    \end{itemize}
    \item What information do you hope to get from benchmarks?
    \begin{itemize}
        \item How would this information be helpful?
        \item What choices does this information inform?
    \end{itemize}
    \item How often do you use benchmarks?
    \begin{itemize}
        \item Would you like to be using benchmarks more or less often? Why?
    \end{itemize}
\end{enumerate}

\textbf{Benchmark Selection and Evaluation}
\begin{enumerate}
    \item Can you describe your process for finding a suitable benchmark?
    \begin{itemize}
        \item What factors are you considering when choosing a benchmark?
        \item When do you consider having looked at 'enough' benchmarks to make an informed decision?
        \item Where do you find these benchmarks?
    \end{itemize}
    \item What is a benchmark that you recently used, and how did you choose it?
    \begin{itemize}
        \item Why did you choose this benchmark?
    \end{itemize}
    \item What benchmarks are you currently using?
    \begin{itemize}
        \item What kinds of decisions have you used them for?
        \item Which of these purposes would you use those benchmarks for: evaluating bias, evaluating performance, evaluating safety?
    \end{itemize}
    \item Where do you find information about benchmarking results?
    \begin{itemize}
        \item How often do you check this information?
        \item How do you use this information?
    \end{itemize}
\end{enumerate}

\textbf{Benchmark Effectiveness and Limitations}
\begin{enumerate}
    \item On a scale from 1-5, how helpful are benchmarks to your desired use?
    \begin{itemize}
        \item Can you tell me more about your rating?
        \item Would any particular changes make them more useful? Less useful?
    \end{itemize}
    \item In which areas do you feel like there exist 'enough' benchmarks vs. which tasks/capabilities do you feel lack a good benchmark for your use case?
    \begin{itemize}
        \item Which of those areas do you think is most important to develop?
        \item Do you have a hypothesis for why there might currently be a lack of benchmarks in those areas?
    \end{itemize}
    \item Do you feel like you're getting enough context information for benchmark results or to use benchmarks?
    \begin{itemize}
        \item If not, what kinds of additional information would be most helpful?
    \end{itemize}
    \item How have benchmarks impacted the outcome of a project or use case?
    \begin{itemize}
        \item Can you talk a bit about the long-term effects of decisions made based on benchmark results?
    \end{itemize}
    \item Can you describe any notable differences you have observed between a model's performance on benchmarks and its performance in real-world applications?
    \begin{itemize}
        \item How did these observations influence your approach to using benchmarks?
        \item How do you think about the tradeoffs between more general benchmarks and those specific to particular use cases?
    \end{itemize}
\end{enumerate}

\textbf{Risks and Rewards of Benchmarking}
\begin{enumerate}
    \item What do you think about the potential risks of using benchmarks for decision making? What about the rewards?
    \begin{itemize}
        \item Have you observed any of these risks or rewards in the context of your work?
        \item Are these risks mitigated enough by benchmark developers/users, in your opinion?
    \end{itemize}
    \item Can you think of good ways to mitigate risks? Are these being implemented by benchmark developers/users?
    \begin{itemize}
        \item Can you say a bit more about how specific strategies might be helpful in mitigating particular risks?
    \end{itemize}
\end{enumerate}

\textbf{Benchmark Improvement and Future Directions}
\begin{enumerate}
    \item Is there anything that you feel the benchmarks you use are lacking?
    \begin{itemize}
        \item If so, what? How would having this thing be helpful to your use case?
    \end{itemize}
    \item How has your use of benchmarks changed over the past few years?
    \begin{itemize}
        \item Have you noticed any trends in how your field approaches/uses benchmarking?
        \item What caused your use to shift?
    \end{itemize}
    \item Are you aware of any collaborative efforts in developing benchmarks in your field?
    \begin{itemize}
        \item What do you think of these collaborative efforts?
    \end{itemize}
    \item In your opinion, what are the traits of an ideal benchmark? Describe the benchmark that would be most useful to your work.
    \begin{itemize}
        \item Can you tell me a bit more about why specific features would be helpful?
    \end{itemize}
\end{enumerate}

\textbf{Closing Questions}
\begin{enumerate}
    \item What are your biggest pet peeves or complaints with benchmarks?
    \item Is there anything else I should know about benchmarks or your use of benchmarks?
    \item Is there anyone else you think we should talk to about this topic?
\end{enumerate}

\end{document}